\documentclass[10pt,twocolumn,letterpaper]{article}

\usepackage{iccv}
\usepackage{epsfig}
\usepackage{graphicx}
\usepackage{amsmath}
\usepackage{amssymb}
\newcommand*{\method}{COMIC}
\usepackage{booktabs}
\usepackage[linesnumbered,ruled,vlined]{algorithm2e}
\usepackage{hyperref}
\usepackage{url}
\usepackage[utf8]{inputenc} 
\usepackage[T1]{fontenc}    
\usepackage{amsfonts}       
\usepackage{nicefrac}       
\usepackage{microtype}      
\usepackage{xcolor}         
\usepackage{wrapfig}
\usepackage{multirow}
\usepackage{dsfont}
\usepackage{setspace}
\usepackage{caption}
\usepackage{amssymb}
\usepackage{subfigure}
\usepackage{fontawesome}
\usepackage{bbm}
\usepackage{colortbl}
\definecolor{myyellow}{rgb}{1,1, 0.6}
\definecolor{myorange}{rgb}{1, 0.8, 0.6}
\definecolor{myred}{rgb}{1, 0.6, 0.6}

\iccvfinalcopy 

\ificcvfinal\fi
\begin{document}

\title{Learning in Imperfect Environment:  Multi-Label Classification \\ with Long-Tailed Distribution and Partial Labels}
\author{Wenqiao Zhang\\
National University of Singapore\\
{\tt\small wenqiao@nus.edu.sg}
\and
Changshuo Liu\\
National University of Singapore\\
{\tt\small changshuo@u.nus.edu}
\and
Lingze Zeng\\
National University of Singapore\\
{\tt\small lingze@nus.edu.sg}
\and
Beng Chin Ooi\\
National University of Singapore\\
{\tt\small ooibc@comp.nus.edu.sg}
\and
Siliang Tang \\
Zhejiang University\\
{\tt\small siliang@zju.edu.cn}
\and
Yueting Zhuang \\
Zhejiang University\\
{\tt\small yzhuang@zju.edu.cn}
}

\maketitle
\ificcvfinal\thispagestyle{empty}\fi

\begin{abstract}
\label{sec:abstract}
Conventional multi-label classification (MLC) methods assume that all samples are fully labeled and identically distributed.  Unfortunately, this assumption is unrealistic in large-scale MLC data 
that has long-tailed (LT) distribution and partial labels (PL).
To address the problem, we introduce a novel task, Partial labeling and Long-Tailed Multi-Label Classification (PLT-MLC), to jointly consider the above two imperfect learning environments.  Not surprisingly, we find that most LT-MLC and PL-MLC approaches fail to solve the PLT-MLC, resulting in significant performance degradation on the two proposed PLT-MLC benchmarks. Therefore, we propose an end-to-end learning framework: \textbf{CO}rrection $\rightarrow$ \textbf{M}odificat\textbf{I}on $\rightarrow$ balan\textbf{C}e, abbreviated as \textbf{\method{}}.  
Our bootstrapping philosophy is to simultaneously correct the missing labels (Correction) with convinced prediction confidence over a class-aware threshold and to learn from these recall labels during training. 
We next propose a novel multi-focal modifier loss that simultaneously addresses head-tail imbalance and positive-negative imbalance to adaptively modify the attention to different samples (Modification) under the LT class distribution.
In addition, we develop a balanced training strategy by distilling the model's learning effect from head and tail samples,  and thus design a balanced classifier (Balance) conditioned on the head and tail learning effect to maintain stable performance for all samples. 
Our experimental study shows that the proposed \method{} significantly outperforms general MLC, LT-MLC and PL-MLC methods in terms of effectiveness and robustness on our newly created PLT-MLC datasets.
\end{abstract}
\section{Introduction}
\label{sec:introduction}
\begin{figure*}[t]
\includegraphics[width=0.9\textwidth]{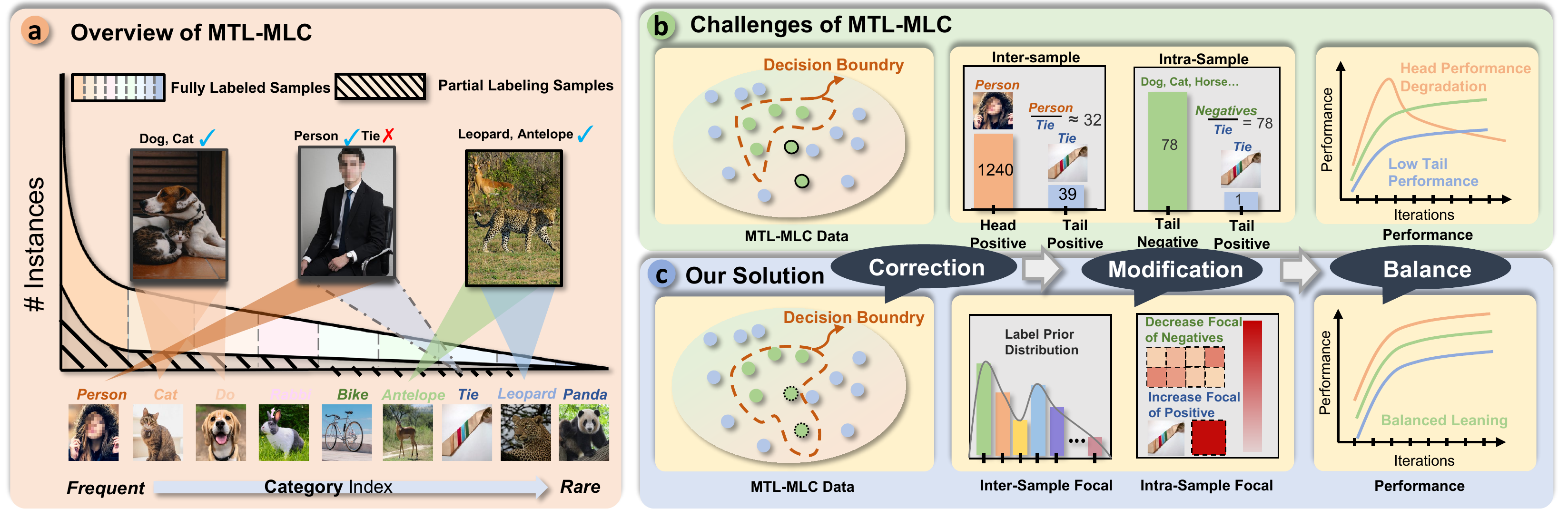}
\centering\caption{(a) illustrates an overview of the proposed PLT-MLC task. 
(b) depicts three key challenges of a PLT-MLC task.  
(c) depicts a concise version 
of
our proposed model for facilitating the PLT-MLC. (\includegraphics[height=0.12in, width=0.12in]{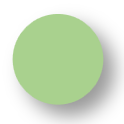} : \emph{positive}, \includegraphics[height=0.12in, width=0.12in]{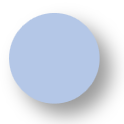} : \emph{negative}, 
\includegraphics[height=0.12in, width=0.12in]{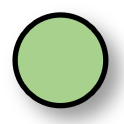} : false \emph{negative},
\includegraphics[height=0.12in, width=0.12in]{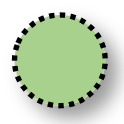} :  corrected \emph{positive})}
\vspace{-0.2cm}
\label{fig:intro}
\end{figure*}
\vspace{-0.2cm}
Images typically contain multiple objects and concepts, highlighting the importance of multi-label classification (MLC)~\cite{tsoumakas2007multi} for real-world tasks.
Along with the wide adoption of deep learning, recent MLC approaches have made remarkable progress in visual recognition~\cite{wehrmann2018hierarchical,yang2018sgm}, but the performance is limited by two common assumptions: 
\emph{all categories have comparable numbers of instances}
and \emph{each training instance has been fully annotated with all the relevant labels}. 
While this conventional setting provides a perfect training environment for various studies, it conceals a number of complexities that typically arise in real-world applications: \textbf{i}) \textbf{Long-Tailed (LT) Class Distribution.} 
With the growth of digital data, the crux of making a large-scale dataset is no longer about where to collect, but how to balance it~\cite{tang2020long}. 
However, the cost of expanding the dataset to a larger class vocabulary with balanced data is not linear — but exponential — since the data is inevitably long-tailed following Zipf’s distribution~\cite{reed2001pareto}.
\textbf{ii}) \textbf{Partial Labels (PL) of Instances.} 
In the case of a large number of categories, it is difficult and even impractical to fully annotate all relevant labels for each image~\cite{yu2014large,zhu2018multi,zhang2022boostmis}. Intuitively,
humans tend to focus on different aspects of image contents
due to \emph{human bias}, \emph{i.e.,}, their preference, personality and sentiment~\cite{zhang2022magic}, 
which indirectly affects how and what we annotate. In fact, LT and PL are often co-occurring, and therefore,  the MLC model must be sufficiently robust to handle different data distributions and imperfect datasets.

In this paper, we present a new challenge for MLC at scale,
Partial labeling and Long-Tailed Multi-Label Classification (PLT-MLC), with
concomitant existence of both PL setting~\cite{yu2014large} and LT distribution~\cite{tang2020long} problems.
As captured in
the overview of PLT-MLC in
Figure~\ref{fig:intro} (a), it has the following three challenges: \textbf{i) False Negative Training.} Under the PL setting, the MLC model treats the un-annotated labels (\includegraphics[height=0.12in, width=0.12in]{figure/false_negative.jpg}) as \emph{negatives} (\includegraphics[height=0.12in, width=0.12in]{figure/negative.jpg}), which may produce sub-optimal decision boundary as it adds noise of false \emph{negative} labels (Figure~\ref{fig:intro} (b)). 
The situation is further exacerbated in the LT class distribution as some \emph{tail} categories are prone to missing annotations in practice. 
For instance, in Figure~\ref{fig:intro} (a), ``\texttt{person}'' is the \emph{head} class in the PLT-MLC dataset and is often notable in an image to labeling for annotators. In contrast, the ``\texttt{tie}''  often occupies a tiny region in the scene compared with the ``\texttt{person}''. The annotator may miss the ``\texttt{tie}'' object, which will aggravate the LT distribution and further increase the difficulty of learning from \emph{tail} classes. 
 \textbf{ii) Head-Tail and Positive-Negative Imbalance.} There are two imbalance issues in a PLT-MLC task: inter-instance  \emph{head}-\emph{tail} imbalance and intra-instance \emph{positive}-\emph{negative} imbalance. As shown in Figure~\ref{fig:intro} (b), the inter-instance ratio of \emph{head positive} (\includegraphics[height=0.12in, width=0.12in]{figure/positive.jpg})  ``\texttt{person}'' (\includegraphics[height=0.12in, width=0.12in]{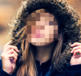}) $:$  \emph{tail positive} (\includegraphics[height=0.12in, width=0.12in]{figure/positive.jpg}) ``\texttt{tie}'' (\includegraphics[height=0.12in, width=0.12in]{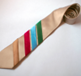}) $\approx 32$ under the LT data distribution, and the intra-instance ratio of  \emph{tail negative} (\includegraphics[height=0.12in, width=0.12in]{figure/negative.jpg}) categories $:$   \emph{tail positive} (\includegraphics[height=0.12in, width=0.12in]{figure/positive.jpg})``\texttt{tie}'' (\includegraphics[height=0.12in, width=0.12in]{figure/tie.jpg}) $ =78 $ as
 an image only contains a small fraction of the \emph{positive} labels. 
 Consequently, a robust PLT-MLC model should address the co-occurring imbalances simultaneously.
 \textbf{iii) Head Overfitting and Tail Underfitting.} Different from the general LT distribution, the classification model downplays the minor \emph{tail} and overplays the major \emph{head}. 
 The PLT-MLC has an extreme LT distribution and Figure~\ref{fig:intro}(c) illustrates an interesting phenomenon of MLC model learning: the general classification model is prone to overfitting to \emph{head} class with extensive samples  and underfitting to \emph{tail} classes with  a few samples. This figure also indicates that only the \emph{medium} class shows a steady growth in performance,  which means that existing LT methods focusing on lifting up the \emph{tail} performance may not solve the PLT-MLC problem satisfactorily.

Suppose a trained model is used to correct the missing labels and then an LT classifier is trained using the updated labels, we might not be able to obtain a satisfying PLT-MLC performance, either. 
While machine learning methods can easily detect the \emph{head} samples, they may have difficulty in identifying the \emph{tail} samples. As a result, the corrected labels may still exhibit an LT distribution that inevitably hurts balanced learning.
 Moreover, even when a general LT classifier affords the trade-off to improve the \emph{tail} performance conditioned on the \emph{head} performance drop, it is still incapable of simultaneously addressing the issue of \emph{head} overfitting and \emph{tail} underfitting problem.
Further,
the decoupled learning paradigm is impractical since it needs the ``stop'' training and human ``re-start'' training, \emph{i.e.}, an end-to-end learning scheme is more desirable.
Thus, these limitations motivate us to reconsider the solution for the PLT-MLC task.

To this end, we propose an end-to-end PLT-MLC framework: \textbf{CO}rrection $\rightarrow$ \textbf{M}odificat\textbf{I}on $\rightarrow$ balan\textbf{C}e (Figure~\ref{fig:intro}), called \textbf{\method{}}, which progressively addresses the three key PLT-MLC challenges.
\textbf{Step 1:} The \emph{Correction} module aims to gradually correct the missing labels according to the predicted confidence and dynamically adjusts the classified loss of the corrected samples under the real-time estimated class distribution.
\textbf{Step 2:} After the label correction, the \emph{Modification} module introduces a  novel Multi-Focal
Modifier (MFM) Loss, 
which contains two focal factors to address the two imbalance issues in PLT-MLC independently. 
Motivated by~\cite{ben2020asymmetric}, the first is an intra-instance \emph{positive}-\emph{negative} factor that determines the concentration of learning on hard \emph{negatives} and \emph{positives} with different exponential decay factors. The second is an inter-instance \emph{head}-\emph{tail} factor that increases the impact of rare categories, ensuring that the loss contribution of rare samples will not be overwhelmed by frequent ones.
\textbf{Step 3:} Finally, the \emph{Balance} module measures the model's optimization direction with a calculated moving average vector of the gradient over all past samples. And thus, we devise a \emph{head} model and a \emph{tail} model by subtracting or adding this moving vector, which can respectively improve \emph{head} and \emph{tail} performance.
Subsequently, a balanced classifier deduces a balanced learning effect under the supervision of the \emph{head} classifier and \emph{tail} classifier. It protects the model training from being too \emph{medium biased}, and hence the balanced classifier is able to achieve the balanced learning schema.
Notably, our solution is an end-to-end learning framework, which is re-training-free and effectively enables balanced prediction.

Our contributions are three-fold: (1) We present a new challenging task: Partial labeling and Long-Tailed Multi-Label Classification (PLT-MLC), together with two newly designed benchmarks: PLT-COCO and PLT-VOC. 
(2) We propose an end-to-end PLT-MLC learning framework, called \method{},  
to effectively perform the PLT-MLC task as a progressive learning paradigm, $i.e.$, \emph{Correction} $\rightarrow$ \emph{Modification} $\rightarrow$ \emph{Balance}. 
(3) Through an extensive experimental study, we show that our method improves all the prevalent LT and ML line-ups on PLT-MLC benchmarks by a large margin. 

\section{Related Works}
\label{sec:related_work}
\noindent\textbf{Long-Tailed MLC.} Deep neural networks excel at learning from large labeled datasets in computer vision~\cite{li2020unsupervised,liu2021swin,han2022survey,zhang2019frame,li2022hero,zhang2021consensus,li2022dilated} and natural language processing tasks~\cite{feng2020language,jawahar2019does,kowsari2019text,lv2023ideal,li2022end,zhang2022magic}. One of the most popular task is Long-Tailed MLC, ~\cite{wu2020distribution} is the first work that addresses the LT-MLC by extending the re-balanced sampling and cost-sensitive re-weighting methods. It proposes an optimized DB Focal method, which does improve the recognition performance
of \emph{tail} classes. Later work, ~\cite{guo2021long} performs uniform and re-balanced samplings from the same training set. Then a two-branch network is developed to enforce the consistency between two branches for collaborative learning on both uniform and re-balanced samplings. However, the above works require careful data initialization, \emph{i.e.}, re-sampling, which is undesirable in practice. Moreover, these methods have not yet considered the missing labeling case, which may not sufficiently deal with the PLT-MLC.

\noindent\textbf{MLC with Partial Labels.} Multi-label tasks often involve incomplete training data, hence several methods have been proposed to solve the problem of multi-label learning with missing labels. A simple solution is to treat the missing labels as \emph{negative} labels~\cite{sun2010multi,li2010optimol,bucak2011multi}. However, performance will drop because a lot of ground-truth \emph{positive} labels are initialized as \emph{negative} labels~\cite{joulin2016learning}. 
Current works on PL-MLC mainly focus on the design of networks and training schemes. The common practice is to utilize the customized networks to learn label correlations or classification confidence to realize correct recognition of missing labels~\cite{durand2019learning,zhang2021simple}. However, the corrected labels learned from a trained model are imbalanced due to the previous training dataset having an LT distribution. Using such recall labels will aggravate the LT distribution in the PLT-MLC dataset and result in an imbalanced performance.

\section{Methodology}
\label{sec:method}
This section describes the proposed PLT-MLC framework. We will present each module and its training strategy.

\begin{figure*}[t]
\includegraphics[width=0.94\textwidth]{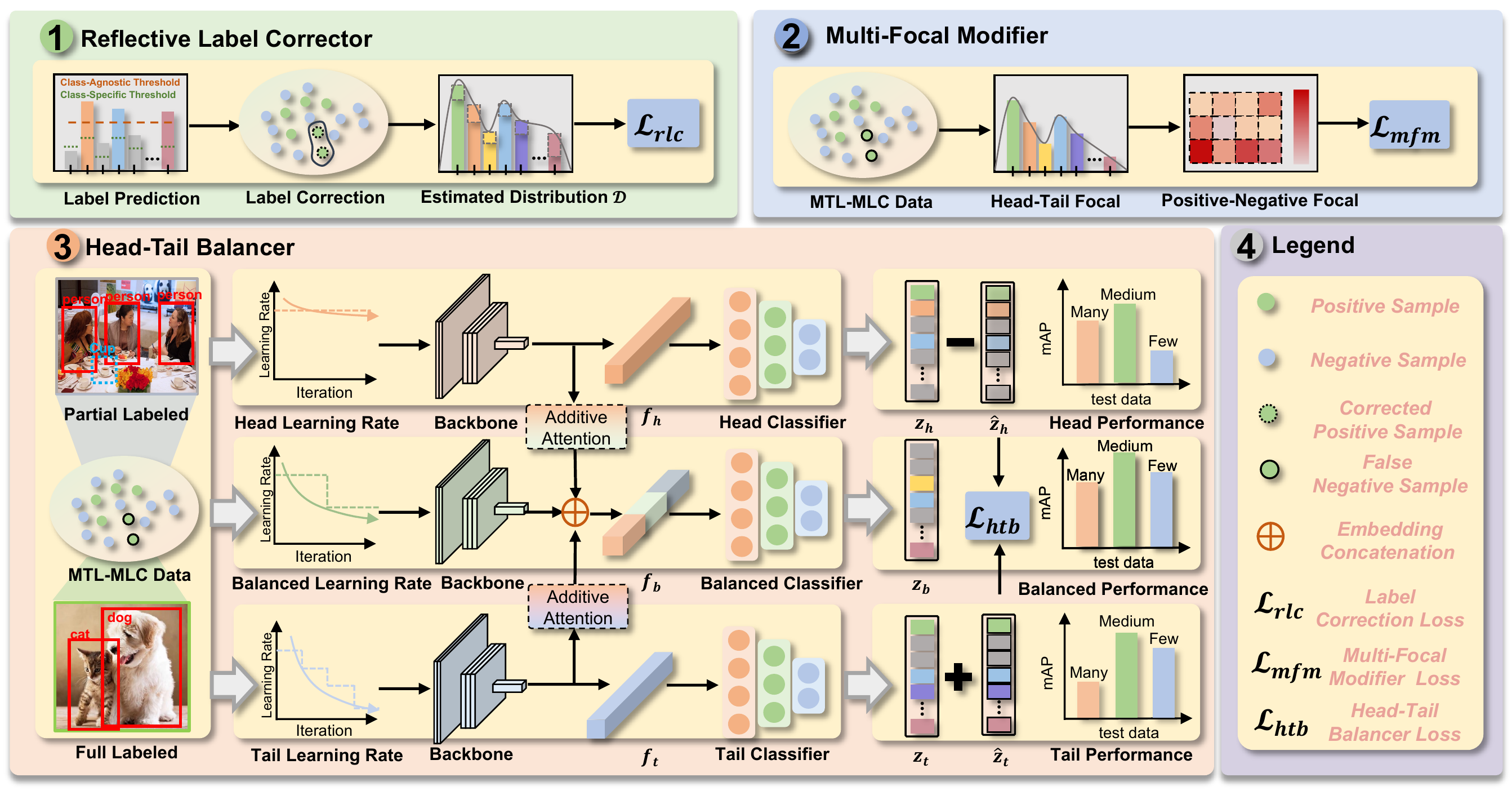}
\centering\caption{\textbf{Overview of \method{}}. RLC module (\emph{Correction}) corrects the missing labels along with the training and dynamically re-weights the sample weight according to the estimated class distribution. MFM module (\emph{Modification}) adjusts the focal of different instances according to \emph{head}-\emph{tail} and \emph{positive}-\emph{negative} imbalance under the extreme LT distribution. HTB module (\emph{Balance}) measures the model's optimization direction and correspondingly develops a balanced learning scheme to produce stable PLT-MLC performance.}
\label{fig:overview}
\end{figure*}

\subsection{Problem Formulation}
Before presenting our method, we first introduce some basic notions and terminologies.
We consider a partially annotated MLC dataset contains $C$ classes and $N$ i.i.d  training samples $\mathcal{S}=\{(\mathcal{I}^{(1)}, y^{(1)}), \cdots, (\mathcal{I}^{(N)}, y^{(N)})\}$, where  $ \mathcal{I}^{(i)}$ denote $i^{th}$ image and label $y^{(i)}$= [$y^{(i)}_1, \cdots, y^{(i)}_c$] $\in \{0, 1\}^{C}$. For a given $i^{th}$ example and category $c$, $y^{(i)}_c$ = 0, 1  respectively means the category is unknown and present.   
Our proposed \method{} solves the PLT-MLC problem in an end-to-end learning manner: \emph{Correction} $\rightarrow$ \emph{Modification} $\rightarrow$ \emph{Balance}, with  \emph{Reflective Label Corrector} (RLC, in Sec.\ref{sec:rlc}), \emph{Multi-Focal Modifier} (MFM, in Sec.\ref{sec:mfm}) and  \emph{Head-Tail Balancer} (HTB, in Sec.\ref{sec:htb}), as illustrated in Figure~\ref{fig:overview}.

These three modules are designed to seek a balanced model $\mathcal{M}_b(\cdot;\Theta_b)$, parameterized by $\Theta_b$, to predict the presence or absence of each class given an input image. We denote 
$p = [p_1, \cdots, p_c]$ as the class prediction, computed by the
model: $p_c = \sigma(z_c)$ where $\sigma$ is the sigmoid function, and
$z_c$ is the output logit corresponding to class $c$. The optimized goal of \method{} can be defined as follows:
\begin{equation}
\begin{aligned}
\underbrace{\mathcal{L}((\mathcal{S});\Theta_b)}_{\rm{\method{}\ Loss}}=  \underbrace{\lambda_c\cdot\mathcal{L}_{rlc}}_{\rm{RLC\ Loss}} +
 \underbrace{\lambda_m\cdot\mathcal{L}_{mfm}}_{\rm{MFM\ Loss}}+ \underbrace{ \lambda_b\cdot\mathcal{L}_{htb}}_{\rm{HTB\ Loss}}
 \label{equ:total}
\end{aligned}
\end{equation}
where $\mathcal{L}_{rlc}$, $\mathcal{L}_{mfm}$ and $\mathcal{L}_{htb}$ denote the loss of RLC, MFM and HTB, respectively. $\lambda_c$, $\lambda_m$ and $\lambda_b$ are hyperparameters.   


\subsection{Reflective Label Corrector}
\label{sec:rlc}
Reflective Label Corrector (RLC) presents a real-time label correction method for missing labels to alleviate the effect of partially labeled samples. The core idea is to examine the \emph{label likelihood} $p$ of each training image and recall the labels with convinced prediction confidence during training.  Interestingly, we found that the model can distinguish a large
number of missing labels with high prediction confidence in
the early training stage, which implies that we can recall these missing labels during training to boost PTL-MLC learning.
Here, we first define a threshold $\tau$ and then check the input sample's label likelihood $p$ to check whether it is greater than $\tau$ and then calculate the average category possibility $P_c$ of past trained data with class $c$. If predicted probabilities $p_c$ are highly confident, \emph{i.e.}, $p_c > {\rm max}\{\tau, P_c\}$, we regard that the sample misses the label of class $c$ and set a pseudo-label $\hat{y}_c$. 
\begin{equation} \hat{y}_c=
\begin{cases} 
 1,  & \mbox{if } p_c>{\rm max}\{\tau, P_c\}, y_c=0\\
0, & \mbox{otherwise}
\end{cases} 
\label{equ:2}
\end{equation}

Thus, the loss of RLC module, \emph{i.e.}, $\mathcal{L}_{rlc}$ utilizes the MFM loss (refer the details in Sec.~\ref{sec:mfm}) with these corrected label $\hat{y}$ for model training:
\begin{equation}
\begin{aligned}
\!\!\!\!\!\!\mathcal{L}_{rlc}(p)\!=\!
\begin{cases} 
 \mathcal{L}^{+}_{mfm}(p),  &\!\!\!\!\mbox{if } \hat{y}=1\\
 \mathop{\mathds{1}}_{(y=1)}  \mathcal{L}^{+}_{mfm}(p)\!+\!\mathop{\mathds{1}}_{(\hat{y}=0)}  \mathcal{L}^{-}_{mfm}(p), & \!\!\!\!\!\mbox{otherwise}
\end{cases} 
\end{aligned}
\label{equ:1}
\end{equation}
where $\mathcal{L}_{mfm}^{+}(p)$ and $\mathcal{L}_{mfm}^{-}(p)$ refer to the generalized MFM loss function for \emph{positives} and \emph{negatives}. Notably, we find that most of the corrected labels belong to \emph{head} class, which may aggravate the LT distribution. To address this issue, we dynamically adjust the inter-sample \emph{head}-\emph{tail} factor $\gamma^{(i)}_{ht}$ (details of this factor are explained in Sec.~\ref{sec:mfm}) according to the dynamic class distribution $\mathcal{D}_t$, which can increase the focal weight for \emph{tail} samples. In addition, we also multiply $\mathcal{L}_{rlc}(p)$ by a  coefficient $\frac{\mathcal{B}_s} {\mathcal{N}_t}$ in each training batch to constrain the loss value, where  $\mathcal{B}_s$ is the batch size and $\mathcal{N}_t$ is the number of corrected labels.

Through this, RLC can gradually and dynamically correct the
potential missing labels during training, which efficiently improves the classifier's performance with recalled labels.

\subsection{Multi-Focal Modifier}
\label{sec:mfm}

We first revisit the focal loss~\cite{lin2017focal}, which is a widely-used solution in the \emph{positive}-\emph{negative} imbalance problem.
It redistributes the loss contribution of easy samples and
hard samples, which greatly weakens the influence of the
majority of \emph{negative} samples. 
\begin{equation} \mathcal{L}_{fl}(p)=
\begin{cases} 
 \mathcal{L}_{fl}^{+}= (1-p)^{\gamma}{\rm log}(p), & {\rm if}\ y=1\\
\mathcal{L}_{fl}^{-}= p^{\gamma}{\rm log}(1-p), &  {\rm if}\ y=0
\end{cases} 
\label{equ:2}
\end{equation}
where $\gamma$ is the focusing parameter, $\gamma$= 0 yields binary cross-entropy. By setting $\gamma$> 0, the contribution of easy \emph{negatives} (with low probability, p $\ll$ 0.5) can be down-weighted in the loss, enabling the model to focus on harder samples.

However, the focal loss may  not satisfactorily resolve the PLT-MLC problem due to two key aspects: 
\begin{itemize}
\item \textbf{Tail Positive Gradient Elimination.} When using focal loss for multi-label training, there is an inner trade-off: high $\gamma$ sufficiently down-weights the contribution from easy \emph{negatives}, but may eliminate the gradients from the  \emph{tail positive} samples~\cite{ben2020asymmetric}. 

\item \textbf{Head-Tail Imbalance.} Imbalance among the \emph{positive} categories also exists in MLC, \emph{i.e.}, \emph{head positive}-\emph{tail positive} imbalance. Rare categories suffer more from severe imbalance issues than frequent ones. 
\end{itemize}

Thus, we propose a  Multi-Focal Modifier (MFM) loss that decouples $\gamma$ at two granularities of focal factors, \emph{i.e.}, an intra-sample \emph{positive}-\emph{negative} (P-N) factor $\gamma_{pn}$ and an inter-sample \emph{head}-\emph{tail} (H-T) factor $\gamma_{ht}$.
\begin{equation} \gamma^{(i)}=
\begin{cases} 
\gamma^{(i)+}=\gamma^{+}_{pn} +w^{+} \cdot \gamma^{(i)}_{ht}, & {\rm if}\ y=1\\
\gamma^{(i)-}=\gamma^{-}_{pn} +w^{-} \cdot \gamma^{(i)}_{ht}, & {\rm if}\ y=0
\end{cases} 
\label{equ:1}
\end{equation}
where $\gamma^{(i)+}$ and $\gamma^{(i)-}$ control the focal of samples with $i^{th}$ class. Similar to~\cite{ben2020asymmetric}, $\gamma^{+}_{pn}$ and $\gamma^{-}_{pn}$ decouple the original decay rate $\gamma$ as two factors, which respectively control the focal of the \emph{positive} and \emph{negative} samples.
Since we are interested in emphasizing the contribution of \emph{positive} samples, we set $\gamma^{-}_{pn}  \geq \gamma^{+}_{pn}$. We achieve better control over the
contribution of \emph{positive} and \emph{negative} samples through the designed loss
function, which assists the network to learn meaningful features
from \emph{positive} samples, despite their rarity.
Another focal factor $\gamma^{(i)}_{ht}$
 is a variable parameter ($\geq 1$) associated with the imbalance degree of the $i^{th}$ class. A bigger value of $\gamma^{(i)}_{ht}$ will increase the weight of \emph{tail} samples to encourage the model to pay more attention to the \emph{positive} \emph{tail} samples, and vice versa.  $w^{+}$ and  $w^{-}$ are the coefficients that adjust the weight at a fine-grained level.  The  $\gamma^{(i)}_{ht}$ is the static class distribution $\mathcal{D}$ of training set with max normalization function $\psi(\cdot)$~\cite{patro2015normalization} to adjust the \emph{head}-\emph{tail} focal. 

After applying the decoupled $\gamma^{(i)+}$ and $\gamma^{(i)-}$ into our MFM loss,
we obtain the loss function as follows (more discussions in the Appendix.):
\begin{equation} \!\!\!\!\mathcal{L}_{mfm}(p)=
\begin{cases} 
 \mathcal{L}_{mfm}^{+}= \sum_{i=1}^{C}  (1-p)^{\gamma^{(i)+}}{\rm log}(p), &{\rm if}\ y=1 \\
\mathcal{L}_{mfm}^{-}=\sum_{i=1}^{C}  p^{\gamma^{(i)-}}{\rm log}(1-p), &{\rm if}\ y=0
\end{cases} 
\label{equ:2}
\end{equation}

By doing so, the MFM module utilizes the multi-grained focal to alleviate the two imbalance problems in the PLT-MLC task, yielding better classification results.

\label{sec:htb}
\subsection{Head-Tail Balancer}
As discussed in the introduction, the extreme LT dataset with numerous \emph{head} samples and a small number  of \emph{tail} samples result in a \emph{head} overfitting and \emph{tail} underfitting learning effect. 
Only the \emph{medium} samples present a superior  performance during training, which fails to obtain the balanced performance for the overall samples.
To address this issue, we develop a balanced strategy that measures the balanced learning effect under the supervision of the \emph{head} classifier and \emph{tail} classifier to achieve balanced results. 
Before we delve into the balanced learning, we first measure the moving average of the gradient in the SGD-based optimizer~\cite{sutskever2013importance}: 
\begin{equation}
\begin{aligned}
\textbf{e}_t=\mu \cdot \textbf{e}_{t-1}+{\rm sum}(g_{t}), \forall t=  1, \cdots, T.
\end{aligned}
\end{equation}
where ${\rm sum}(g_{t})$ is the accumulated gradient at iteration $t$, $\mu$ is the momentum decay. The average moving vector
$\textbf{e}_t$ records the model's optimization tendency by $\textbf{e}_{t-1}$ and ${\rm sum}(g_{t})$. 

In our empirical study, we observe that only the \emph{medium} samples obtain a stable learning effect
from the early to late training stage, mainly due to the extreme LT distribution. 
To simulate the learning effect towards \emph{head}/\emph{tail} samples, as depicted in Figure~\ref{fig:overview}(c), we reduce/add the moving vector $\textbf{e}_t$ at each step in \emph{head} model $\mathcal{M}_h$ and \emph{tail} model $\mathcal{M}_t$ respectively, to assist the balanced model $\mathcal{M}_b$ for balanced learning.  Notably, we set different learning rate decays for each model to further explore the balanced learning effect. The three models are parallel-trained with their own backbone and classifier. In the feature learning stage, we develop an additive attention~\cite{bahdanau2014neural} that computes the relevance of balanced features $\hat{\textbf{f}}_b$, and \emph{head} features $\textbf{f}_h$, \emph{tail} features $\textbf{f}_t$ extracted from corresponding backbones.
\begin{equation}
\begin{aligned}
\textbf{f}_b= {\rm Attn}(\hat{\textbf{f}}_b,[\textbf{f}_h,\textbf{f}_t]))+\hat{\textbf{f}}_b \\
\end{aligned}
\end{equation}
where $Attn(\cdot)$ is the additive attention mechanism.

Then updated ${\textbf{f}}_b$, $\textbf{f}_h$, $\textbf{f}_t$ are input to their classifiers to obtain their logits. We develop the multi-head classifier with normalization~\cite{gidaris2018dynamic,liu2019large,tang2020long}, which has already been embraced by various methods of empirical practice.
The multi-head strategy~\cite{vaswani2017attention} equally divides the channel of weights and features into $N_g$ groups, which can be considered as $N_g$ times of fine-grained sampling. 
\begin{equation}
\begin{aligned}
\textbf{z}_x=\frac{\rho}{N_g}\sum_{k=1}^{N_g}\frac{w_k^{\top}{\textbf{f}_x}}{(||w_k||+\eta)||\textbf{f}_x||}, x \in \{h,t,b\}
\end{aligned}
\end{equation}
where $\rho$ is a  scaling factor akin to the inverse temperature in Gibbs distribution, $\eta$ is a class-agnostic baseline energy. $w_k$ is the $k^{th}$ learned parameter matrix.

\begin{table*}[t]
  \caption{
  \textbf{Performance comparison of the proposed method and baselines on PLT-MLC datasets} (\texttt{PLT-COCO} and \texttt{PLT-VOC}). \textbf{\emph{$E2E^*$}} indicates that the PLT model is learned in an end-to-end manner. A larger score has better performance. Improv. indicates performance improvement.  Acronym notations of baselines can be found in Sec.~\ref{sec:baseline}.
  We color each row as the \colorbox{myred}{\textbf{best}}, \colorbox{myorange}{\textbf{second best}} and  \colorbox{myyellow}{\textbf{lowest score}}.} 
  \label{tab:experiment_rs}
  \centering
 \renewcommand{\arraystretch}{1.2}
 \resizebox{0.98\textwidth}{!}{
{
    \begin{tabular}{c|c|c|c|c|c|c|c|c|c|c}
    \toprule[2pt]
     \multirow{2}{*}{\textbf{Category}}&\multirow{2}{*}{\textbf{Methods}} & \multirow{2}{*}{\textbf{\emph{$E2E^*$}}} & \multicolumn{4}{c|}{$\texttt{PLT-COCO Dataset}$}  & \multicolumn{4}{c}{$\texttt{PLT-VOC Dataset}$} \\
     \cline{4-7}\cline{8-11}
    &
    & & \emph{Many Shot} & \emph{Medium Shot}& \emph{Few Shot}&  \emph{Total Shot} & \emph{Many Shot} & \emph{Medium Shot}& \emph{Few Shot}&  \emph{Total Shot} \\
    \midrule[1pt]
    \midrule[1pt]

    \multirow{4}{*}{MLC} & BCE ~\cite{zhang2018generalized}  &\faCheckCircle 
    & 42.57$\pm$0.11 & \colorbox{myyellow}{56.67$\pm$0.19} & \colorbox{myyellow}{46.40$\pm$0.60} & \colorbox{myyellow}{48.92$\pm$0.23}  &
    67.37$\pm$0.18 & 88.27$\pm$0.39 & 83.79$\pm$0.41 & 78.79$\pm$0.14 \\

     & Focal~\cite{lin2017focal} &\faCheckCircle& \colorbox{myyellow}{41.05$\pm$0.07} & 58.33$\pm$0.12 & 53.58$\pm$0.31 & 51.39$\pm$0.15
     & \colorbox{myyellow}{67.02$\pm$0.11} & 87.49$\pm$0.18 & 82.82$\pm$0.78 & 78.13$\pm$0.23\\

     & ASL~\cite{ben2020asymmetric} &\faCheckCircle & 41.60$\pm$0.17  & 58.15$\pm$0.15 & 52.67$\pm$0.17  & 51.20$\pm$0.08 & 67.67$\pm$0.10 & 87.79$\pm$0.13 & 82.23$\pm$0.55 & 78.35$\pm$0.11\\

    \midrule[1pt]   
    
        \multirow{4}{*}{LT-MLC} & DB~\cite{wu2020distribution} &\faCheckCircle  & 44.83$\pm$0.31 & 58.96$\pm$0.24 & 53.82$\pm$0.47 & 52.16$\pm$0.36 & 69.22$\pm$0.28 & 88.56$\pm$0.42 & 83.72$\pm$0.35 & 78.86$\pm$0.23\\ 

     & DB-Focal~\cite{wu2020distribution} &\faCheckCircle & 45.76$\pm$0.25 & \colorbox{myorange}{59.74$\pm$0.21} & 53.85$\pm$0.16 & 52.57$\pm$0.27 & 68.96$\pm$0.22 & \colorbox{myorange}{88.89$\pm$0.18} & 83.42$\pm$0.20 & 78.90$\pm$0.26\\  
     & LWS~\cite{kang2019decoupling} &- & 44.86$\pm$0.58& 58.79$\pm$0.63& 53.48$\pm$0.51& 52.86$\pm$0.60 & 69.08$\pm$0.44& 88.24$\pm$0.55& 83.46$\pm$0.47& 78.28$\pm$0.49\\ 
    \midrule[1pt]   
    
    \multirow{4}{*}{PL-MLC} & Pseudo-Label~\cite{lee2013pseudo} &-& 41.41$\pm$0.41 & 57.46$\pm$0.35 & 53.12$\pm$0.33 & 51.67$\pm$0.37 & 67.38$\pm$0.24 & 87.58$\pm$0.35 &83.26$\pm$0.42 &78.32$\pm$0.30 \\ 

     & ML-GCN~\cite{chen2019multi} &\faCheckCircle&43.43$\pm$0.53 & 58.46$\pm$0.61 & 53.74$\pm$0.48 & 52.14$\pm$0.55 & 68.46$\pm$0.44 & 88.17$\pm$0.61 &82.46$\pm$0.38 &79.02$\pm$0.56 \\ 
     
     & Hill~\cite{zhang2021simple}  &\faCheckCircle  & 42.50$\pm$0.16 & 56.89$\pm$0.19 & 47.31$\pm$0.37 & 49.28$\pm$0.09& 
     68.79$\pm$0.15 & \colorbox{myyellow}{86.70$\pm$0.17} & \colorbox{myyellow}{78.15$\pm$0.99} & \colorbox{myyellow}{77.40$\pm$0.22} \\ 
     & P-ASL~\cite{ben2022multi} &\faCheckCircle & 43.09$\pm$0.05& 57.67$\pm$0.07 & 53.46$\pm$0.22 & 51.75$\pm$0.17&68.95$\pm$0.22& 87.24$\pm$0.13 & 83.37$\pm$0.33 & 78.96$\pm$0.16 \\ 
 \midrule[1pt]
    \midrule[1pt]
     \multirow{3}{*}{PLT-MLC}& 
       Head Model (Ours)&\faCheckCircle&\colorbox{myorange}{47.59$\pm$0.09} & 59.07$\pm$0.12 & 52.35$\pm$0.28 & \colorbox{myorange}{53.30$\pm$0.19} & \colorbox{myorange}{72.91$\pm$0.28}& 88.59$\pm$0.31&  82.12$\pm$0.27& \colorbox{myorange}{80.70$\pm$0.30} \\ 
     & Tail Model (Ours)&\faCheckCircle&46.30$\pm$0.25 & 58.76$\pm$0.29 & 53.38$\pm$0.14 & 53.09$\pm$0.27 &71.65$\pm$0.34& 88.68$\pm$0.41& 83.51$\pm$0.24& 80.58$\pm$0.36\\ 
     
     \rowcolor{gray!40}  & \textbf{\method{} (Ours)}   &\faCheckCircle & \colorbox{myred}{\textbf{49.21}$\pm$\textbf{0.22}}& \colorbox{myred}{\textbf{60.08}$\pm$\textbf{0.13}} & \colorbox{myred}{\textbf{55.36}$\pm$\textbf{0.21}}  & \colorbox{myred}{\textbf{55.08}$\pm$\textbf{0.14}} & \colorbox{myred}{\textbf{73.10}$\pm$\textbf{0.35}}&\colorbox{myred}{\textbf{89.18}$\pm$\textbf{0.45}}&\colorbox{myorange}{\textbf{84.53}$\pm$\textbf{0.48}} & \colorbox{myred}{\textbf{81.53}$\pm$\textbf{0.35}}\\
    \bottomrule[2pt]
    \end{tabular}
    }
}
\label{tab:main}
\end{table*}

Subsequently, we measure the \emph{head} and \emph{tail} learning effect by subtracting and adding the average moving vector $\textbf{e}_t$ to the logits of \emph{head} model and \emph{tail} model, respectively:
\begin{equation}
\begin{aligned}
\!\!\!\!\hat{\textbf{z}}_x=\textbf{z}_x \pm \frac{\rho}{N_g}\sum_{k=1}^{N_g}\frac{{\rm sim}(\textbf{z}_x,\textbf{e}_t)\cdot(w_j)^{\top}\textbf{e}_t}{||w_k||+\eta}, x \in \{h,t\}
\end{aligned}
\end{equation}
where $sim(\cdot,\cdot)$ measures the cosine similarity of vectors.

After obtaining the logits of $\hat{\textbf{z}}_h$, $\hat{\textbf{z}}_t$ and $\textbf{z}_b$, 
the balanced learning effect needs to distill the \emph{head} and \emph{tail} knowledge from $\hat{\textbf{z}}_h$ and $\hat{\textbf{z}}_t$ to enable the stable learning for all samples. Hence, we develop the \emph{head}-\emph{tail} loss:
\begin{equation}
\begin{aligned}
\!\!\!\mathcal{L}_{htb} &= \kappa_h \cdot \mathcal{L} (\phi(\hat{\textbf{z}_h}) \cdot 
\phi({\textbf{z}_{b}}))+\kappa_t\cdot \mathcal{L} (\phi(\hat{\textbf{z}_t}) \cdot \phi({\textbf{z}_b}))\!\!\!\!
\end{aligned}
\end{equation}
where $\mathcal{L}$ is $\mathcal{L}_{mfm}$ and $\phi(\cdot)$ is softmax function. $\kappa_h$ and $\kappa_t$ are adaptive weights for \emph{head} and \emph{tail} learning that calculated by $\kappa_h=\frac{({\mathcal{L}(\hat{z_h}))}^{\alpha}}{{(\mathcal{L}(\hat{z_t}))}^{\alpha} + {(\mathcal{L}(\hat{z_h}))}^{\alpha}}$ and $\kappa_t=\frac{({\mathcal{L}(\hat{z_t}))}^{\alpha}}{{(\mathcal{L}(\hat{z_t}))}^{\alpha} + {(\mathcal{L}(\hat{z_h}))}^{\alpha}}$, respectively.  $\alpha$ is a scaling factor and we study its effect in Sec.~\ref{sec:hyp}. Such loss can be regarded as the empirical risk minimization (ERM)~\cite{donini2018empirical}, which adaptively distills the knowledge from the \emph{head} and \emph{tail} models,  enabling
the balanced model is not biased to \emph{medium} samples and produces a balanced learning effect for the PLT-MLC task.

\section{Experiments}
We verify \method{}'s effectiveness on  two proposed PLT-MLC datasets and then discuss \method{} ’s properties with controlled studies. 
\begin{table}[t]
  \caption{\textbf{Ablation study of different modules.} M,C,B represent correction, modification and balance learning, respectively. }
  \label{tab:ablation_module_study}
  \centering
 \renewcommand{\arraystretch}{1.2}
 \resizebox{0.48\textwidth}{!}{
{
    \begin{tabular}{r|ccc|c|c|c}
    \toprule[1pt]
    \multirow{2}{*}{Models} & \multicolumn{3}{c|}{{\textbf{Setting}}} & \multicolumn{3}{c}{\texttt{PLT-COCO Dataset}}\\
    \cline{2-7}
    & M&C&B& \emph{Total mAP} & \emph{Average mAP} & \emph{Recall} \\
    \midrule[1pt]
    \midrule[1pt]
    -\textbf{RLC} & &\faCheckCircle &\faCheckCircle & 54.70$\pm$0.13 & 54.42$\pm$0.15 & 85.26$\pm$0.08  \\
    -\textbf{MFM} &\faCheckCircle  & &\faCheckCircle & 54.60$\pm$0.13 & 54.33$\pm$0.13 & 84.59$\pm$0.19 \\
    -\textbf{HTB} &\faCheckCircle  &\faCheckCircle & & 53.65$\pm$0.31 & 53.36$\pm$0.31 & 84.19$\pm$0.23 \\
    \midrule[1pt]
    \textbf{\method{}} &\faCheckCircle  &\faCheckCircle &\faCheckCircle & \textbf{55.08}$\pm$\textbf{0.14} & \textbf{54.88}$\pm$\textbf{0.19} & \textbf{88.19}$\pm$\textbf{0.22} \\
    \bottomrule[1pt]
    \end{tabular}
    }
}
\label{tab:aba}
\end{table}
\subsection{Experimental Setup}
\noindent$\textbf{Dataset Construction.}$  
The proposed method is analyzed on 
the created LT versions of two MLC benchmarks (COCO~\cite{lin2014microsoft} and VOC~\cite{everingham2010pascal}), called $\texttt{PLT-COCO}$ and $\texttt{PLT-VOC}$, respectively. The missing rate of $\texttt{PLT-COCO}$ is 40\%  and it contains 2,962 images from 80 classes. 
The maximum training number for each class is 1,240 and the minimum number is 6.
We select 5000 images from the test set of COCO2017 for evaluation. $\texttt{PLT-VOC}$ has the same missing rate setting and contains 2,569 images from 20 classes, in which 
the maximum training number for each class is 1,117 and the minimum number is 7.
We evaluate the performance on VOC2007 test set with 4,952 images. More details about the dataset construction can be found in the Appendix.

\noindent$\textbf{Implementation Details.}$  
We employ the ResNet-50~\cite{he2016deep} as the backbone model to conduct the PLT-MLC task. We train our model with a standard Adam~\cite{kingma2014adam} optimizer in all the experiments. The images will be randomly cropped and resized
to 224 × 224 together with standard data augmentation.
Besides, we use an identical set of hyperparameters ($B$=32, $Mo$=0.9, $E_{max}$=40 )\footnote{$B$ and $Mo$  refer to the batch size and  momentum in the Adam. } across all the datasets. More details of implementation are in Appendix.

\noindent \textbf{Evaluation Metrics.} Following~\cite{wu2020distribution}, we split these classes into three groups according to the
number of their training examples: each \emph{head} class contains over 100 samples as a many shot,
each \emph{medium} class has 20 to 100 samples as a medium shot, and each \emph{tail} class has
less than 20 samples as a low shot. The total shot indicates all the test samples. 
We evaluate mean average precision (\emph{mAP}) for all the classes and recall for missing label settings.

\noindent\textbf{Comparison of the  Methods.}
\label{sec:baseline}
To quantify the efficacy of the proposed framework, we use several baselines for performance comparison according to different aspects\footnote{We only compare with the approaches that have open source code.}. MLC methods: BCE~\cite{zhang2018generalized}, Focal~\cite{lin2017focal}, ASL~\cite{ben2020asymmetric}.
LT-MLC methods: DB~\cite{wu2020distribution}, DB-Focal~\cite{wu2020distribution} and LWS~\cite{kang2019decoupling}.
PL-MLC methods: Hill~\cite{zhang2021simple}, Pseudo-Label~\cite{lee2013pseudo}, ML-GCN~\cite{chen2019multi} and P-ASL~\cite{ben2022multi}.

\begin{table}[t]
  \caption{\textbf{Performance comparison under different missing labeled settings.} $0\%$ indicates an LT dataset that is fully labeled.}
  \label{tab:experiment_to_be_finished_1}
  \centering
 \renewcommand{\arraystretch}{1.2}
 \resizebox{0.48\textwidth}{!}{
{
    \begin{tabular}{c|c|c|c|c}
    \toprule[1pt]
    \multirow{2}{*}{{Missing Ratio}} &\multicolumn{4}{c}{\texttt{PLT-COCO Dataset}} \\
     \cline{2-5}
     & \emph{Total Shot} & \emph{Many Shot} & \emph{Medium Shot} & \emph{Low Shot} \\
   
    \midrule[1pt]
    \midrule[1pt]
    \textbf{0}\% & 57.07$\pm$0.09 & 52.21$\pm$0.11 & 59.98$\pm$0.12 & 61.12$\pm$0.24 \\
    \textbf{30\%} & 55.80$\pm$0.17 & 49.97$\pm$0.11 & 62.59$\pm$0.15 & 54.56$\pm$0.17  \\
    \textbf{40\%} & 54.75$\pm$0.19 & 48.93$\pm$0.24 & 60.31$\pm$0.21 & 54.14$\pm$0.21  \\
    \textbf{50\%} & 54.69$\pm$0.15 & 48.74$\pm$0.12 & 56.68$\pm$0.16 & 57.25$\pm$0.24  \\
    \bottomrule[1pt]
    \end{tabular}
    }
}
\end{table}
\subsection{Overall Performance}
Table~\ref{tab:main} summarizes the quantitative PLT-MLC results of our framework and baselines on \texttt{PLT-COCO} and  \texttt{PLT-VOC}. We make the following observations: 
1) In general, irrespective of the different shot scenarios, compared to SoTAs, \method{} achieves the best performance on almost all the metrics across both datasets. 
In particular, \method{} outperforms other baselines in terms of total shot's mAP by a large margin (\texttt{PLT-COCO}: \textbf{\underline{1.78\% $\sim$ 6.16\%}},  \texttt{PLT-VOC}: \textbf{\underline{0.83\% $\sim$ 4.13\%}} ) for PLT-MLC task. 
2) Besides, we can observe from Table~\ref{tab:main} that the LT methods outperform PL baselines
in most $X$-$shot$ situations. We believe the underlying reason behind this is
that LT data distribution hurts the classification capability for MLC models more seriously than PL. 
Besides, the label correction may have aggravated the LT issue and further result in 
performance reduction. 
3) Benefiting from the carefully designed HTB module, our \method{} not only achieves the highest total mAP score but also yields balanced results with a narrowed performance gap in different shot metrics. These results demonstrate the superiority of our proposed model.

\begin{figure}[t]
\includegraphics[width=0.5\textwidth]{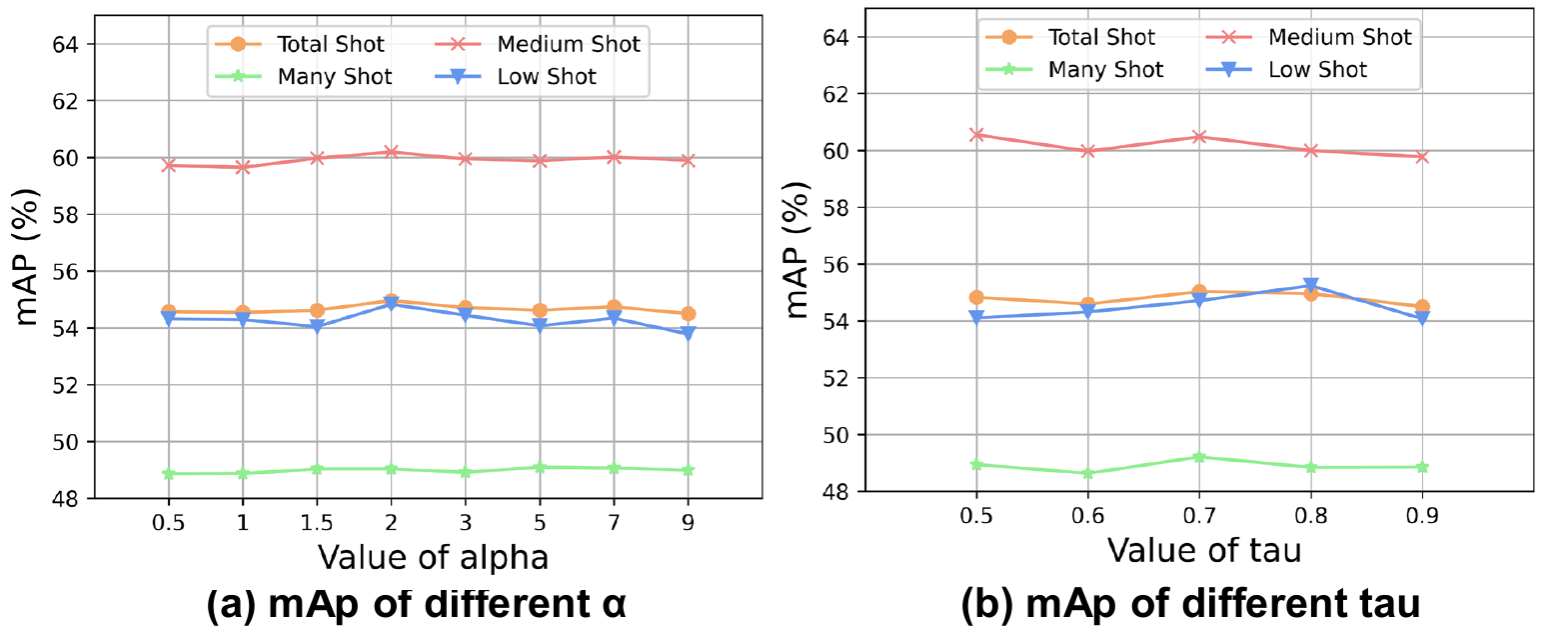}
\centering\caption{\textbf{Ablations with respect to coefficient
$\alpha$ and $\tau$.}}
\label{fig:hyper}
\end{figure}

\begin{figure}[t]
\includegraphics[width=0.5\textwidth]{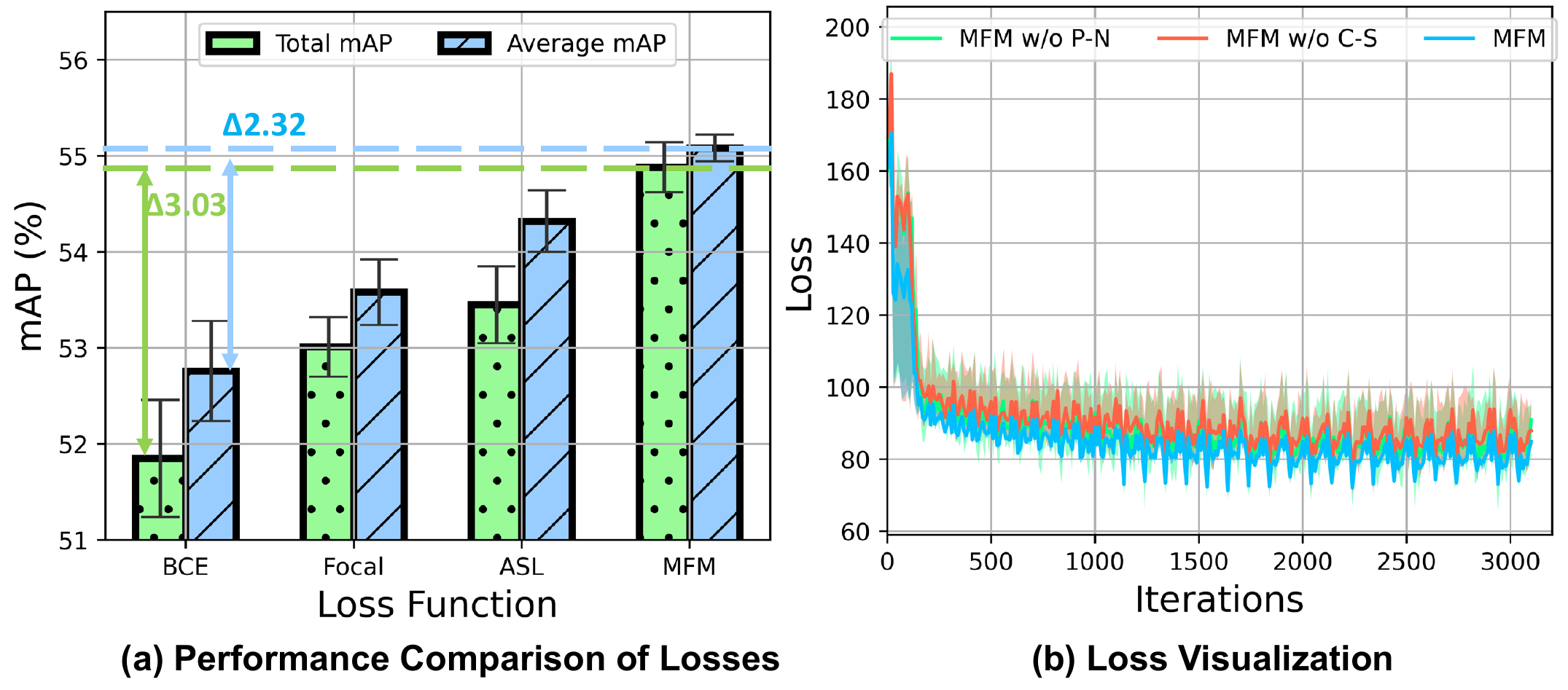}
\centering\caption{\textbf{MLT-MLC results using different losses.}}
\label{fig:losses}
\end{figure}

\subsection{Ablation Study}
\noindent\textbf{Effectiveness of Each Component.} 
We conduct an ablation study to illustrate the effectiveness of each component in Table~\ref{tab:aba}.  Comparing \method{} and \method{}(-RLC) (Row 1 \emph{v.s} Row 4), the label (\underline{\emph{Correction}}) mechanism contributes 0.38\%  improvement on total mAP. 
The results of Row 2 show the mAP improvement of the MFM (\underline{\emph{Modification}}).
Meanwhile, 
Row 3 indicates that it suffers from noticeable performance degradation without the (\underline{\emph{Balance}}) learning. To sum up, we can observe that the improvement of each module is distinguishable. Combining all the
components, our \method{} exhibits steady improvement over the baselines.

\begin{figure}[t]
\includegraphics[width=0.5\textwidth]{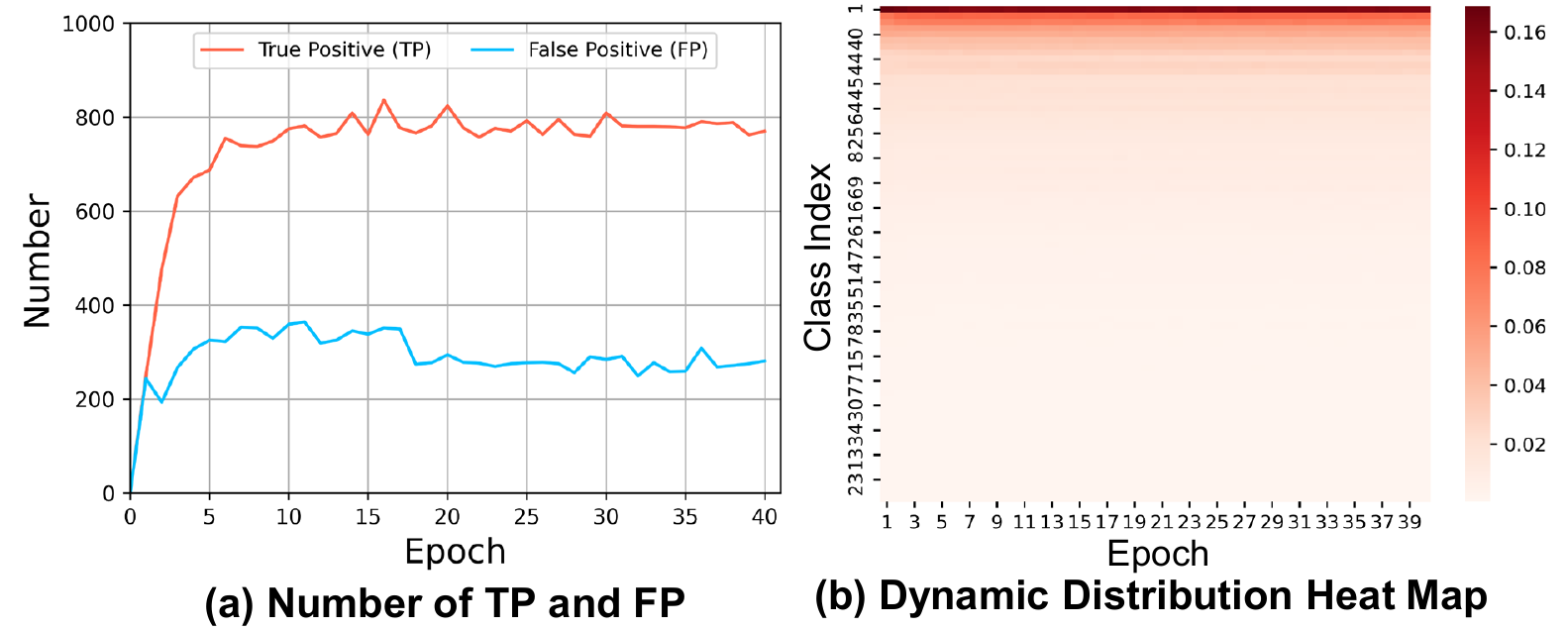}
\centering\caption{\textbf{In-depth analysis of label correction.}}
\label{fig:rlc}
\vspace{-3mm}
\end{figure}

\noindent\textbf{Ablation of Missing Rate.} To study the effect of partial labels that affect \method{}'s results, 
we evaluate the performance under different  missing rates (MR) of labels (from 0\% $\sim$ 50\%). 
Not surprisingly, when the MR decreases, the accuracy of \method{}  increases on all the metrics. 
We also find that the performance gap between different shots is consistently small in all MR settings.   
The results demonstrate the generalizability of the proposed \method{} that it can produce stable and balanced results under different MR settings.

\begin{figure*}[t]
\includegraphics[width=1\textwidth]{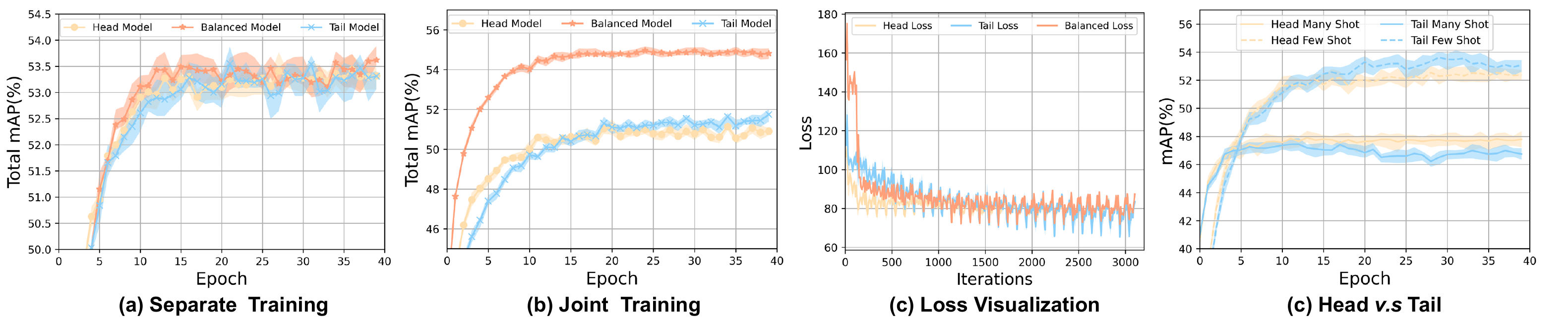}
\centering\caption{\textbf{Analysis of balanced learning of \method{}.} (a) and (b) depict the total mAP of separate and joint training of \method{} within the 40 epochs. (c) summarizes the loss visualization of \emph{head}, balanced and tail models with joint training.  (d) demonstrates \emph{head} and \emph{tail} models respectively 
optimize the \emph{head} and \emph{tail} class's performance. }
\label{fig:balance}
\end{figure*}

\begin{table}[t]
  \caption{\textbf{Ablation of MFM.} $\downarrow$ indicates the mAP decay.}
  \label{tab:experiment_to_be_finished_2}
  \centering
 \renewcommand{\arraystretch}{1.2}
 \resizebox{0.48\textwidth}{!}{
{
    \begin{tabular}{cc|c|c|c|c}
    \toprule[1pt]
    \multicolumn{2}{c|}{{\textbf{MFM Factor}}} & \multicolumn{3}{c}{\texttt{PLT-COCO Dataset}}\\
    \cline{1-2}\cline{3-6}
    P-N & H-T & \emph{Total Shot} & \emph{Many Shot} & \emph{Medium Shot} & \emph{Low Shot}\\
    \midrule[1pt]
    \midrule[1pt]
      &\faCheckCircle  & 54.44 ($\downarrow$ 0.64) & 48.65 ( $\downarrow$ 0.56) & 60.00 ($\downarrow$ 0.08 )&  53.81 ($\downarrow$ 1.55 )\\
         \faCheckCircle& & 53.70  ($\downarrow$ 1.38) & 48.38 ($\downarrow$ 0.83 )& 58.99  ($\downarrow$ 1.09 )&  52.91 ($\downarrow$ 2.45)\\
    \midrule[1pt]
    \faCheckCircle  &\faCheckCircle  & 55.08& 49.21& 60.08 &55.36\\
    \bottomrule[1pt]
    \end{tabular}
    }
}
\label{tab:mfm}
\end{table}

\noindent\textbf{Hyperparameter $\alpha$ and $\tau$.} 
\label{sec:hyp}
We investigate the impact of hyper-parameter $\alpha$ and $\tau$ for the PLT-MLC task. The mAPs 
of different hyper-parameter settings on \texttt{PLT-COCO} are shown in Figure ~\ref{fig:hyper}.  This figure
 suggests that the optimal choices of $\alpha$ and $\tau$ are around
$2$ and $0.7$, respectively. Either increasing or decreasing these values results in performance decay.

\label{sec:experiments}

\subsection{In-Depth Analysis}
We further validate several vital issues of the proposed
 \emph{Correction $\rightarrow$ ModificatIon $\rightarrow$ Balance} learning paradigm  by answering the three questions as follows.

\begin{sloppypar}
\noindent Q1: Can the model trust the recalled labels distinguished by RLC?  To build the insight on the effectiveness of the label correction mechanism in \method{}, we visualize the \emph{true positive} (TP) and \emph{false positive} (FP) in Figure~\ref{fig:rlc}(a)). This figure suggests that the RLC module can distinguish a large number of missing labels with high prediction confidence in the early training stage, meanwhile, sum(TP) $\gg$ sum(FP). However, Figure~\ref{fig:rlc}(b) of corrected samples also reveals LT class distribution with respect to the original training set. 
To address this issue, we dynamically adjust the sample weight conditioned on the real-time distribution  to produce a stable performance. Table~\ref{tab:dyn_sta} reports results under different distributions, which shows that the mAP performance decreases when using static weight conditioned on dynamic distribution, especially for the \emph{tail} samples. 
In contrast, appropriately using these corrected labels with a dynamic sample weight can effectively improve the PLT-MLC performance.
\end{sloppypar}

\begin{sloppypar}
\noindent Q2: How does the MFM module affect the PLT-MLC performance?  Here, we evaluate the effectiveness of the multi-focal modifier (MFM) loss compared with different loss functions.
 Figure~\ref{fig:losses}(a) shows the loss ablation results using different losses in our \method{}. Our developed  MFM  outperforms existing losses, as the designed loss considers the key point of the \emph{head}-\emph{tail}  and \emph{positive}-\emph{negative} imbalance under the extreme LT distribution in the PLT-MLC task. 
There are two components in MFM, which are the \emph{positive}-\emph{negative} (P-N) factor and \emph{head}-\emph{tail} (H-T) factor. To demonstrate the effect of each component, we train the model with the individual factor in the proposed MFM. As shown in Table.~\ref{tab:mfm}, both the P-N factor and the H-T factor play significant roles in MFM. For the H-T factor, it achieves an improvement from 53.7\% mAP to 55.08\% mAP. 
Meanwhile, it brings a significant gain on \emph{tail} categories with 2.45\% mAP improvement, indicating its effectiveness to alleviate the severe \emph{positive}-\emph{negative} imbalance problems in the LT class distribution. 
As for the P-N factor, it brings a steady boost on all shot settings which means it can further alleviate the \emph{positive}-\emph{negative} issue.  Additionally, Figure~\ref{fig:losses}(b) indicates that the MFM loss decreases faster and smoother than the two variants of MFM without different factors, demonstrating its superiority in the PLT-MLC task further. 
\end{sloppypar}

\begin{table}[t]
  \caption{\textbf{Results under static and dynamic distribution. }}
  \label{tab:experiment_to_be_finished_1}
  \centering
 \renewcommand{\arraystretch}{1.2}
 \resizebox{0.48\textwidth}{!}{
{
    \begin{tabular}{c|c|c|c|c}
    \toprule[1pt]
    \multirow{2}{*}{\textbf{Distribution}} &\multicolumn{4}{c}{\texttt{PLT-COCO Dataset}} \\
    \cline{2-5}
     & \emph{Total Shot} & \emph{Many Shot} & \emph{Medium Shot} & \emph{Low Shot} \\
    \cline{1-5}
    \textbf{Static}  & 54.67  (0.41 $\downarrow$) & 48.86 (0.35 $\downarrow$)  & 59.90 (0.18 $\downarrow$)  & 54.41 (1.15 $\downarrow$) \\
    \textbf{Dynamic} & 55.08 & 49.21  & 60.08 & 55.56 \\
    \bottomrule[1pt]
    \end{tabular}
    }
}
\label{tab:dyn_sta}
\end{table}

\begin{sloppypar}
\noindent Q3: How does the HTB module benefit the balanced learning? We systematically present the explicit benefits of the balanced learning strategy in multi-view. 1)  Figure~\ref{fig:balance} (a) and (b) show the comparison between separate and joint training of \emph{head}, balanced and \emph{tail} model
with respect to the total mAP on \texttt{PLT-COCO} dataset. An interesting phenomenon is that the \emph{detached}  \emph{head} and \emph{tail} models slightly outperform the \emph{joint}  \emph{head} and \emph{tail} models but suffer from an unstable performance. In contrast, the accuracy of the \emph{joint} trained balanced model increases much faster and smoother than the  \emph{detached} balanced model
which also yields a stable performance and faster convergence speed. This phenomenon is reasonable as the main optimization objective in joint training is to improve the balanced model's performance.
It can be regarded as the knowledge distillation effect that 
 enables the balanced model to learn from the \emph{head biased}  and \emph{tail biased} model, and this in turn facilitates the PLT-MLC learning.
 2) During the competition of \emph{head} \emph{v.s} \emph{tail}, the \emph{head} model's loss drops faster (shown in Figure~\ref{fig:balance} (c)) and is biased to optimizing the \emph{head} samples, while the \emph{tail} model produces an opposite result. Such \emph{head} and \emph{tail} biased results form a foundation that enables our \method{} to be trained in a balanced and stable learning effect. 
 3) We also perform the analysis of the different balanced learning blocks in our \method{}. As presented in Table 6 (in the Appendix), the DL, NC and AMV contribute 0.05\%, 1.08\% and 0.37\% improvement on total shot mAP.  
 The observations and analysis verify the
effectiveness of balanced learning for being able to study from the \emph{head} and \emph{tail} models, thereby achieving the PLT-MLC improvement.

\end{sloppypar}
\section{Conclusions}
\label{sec:conclusion}
We have presented a fire-new task called PLT-MLC and correspondingly developed a novel framework, named \method{}.
\method{} simultaneously addresses the partial labeling and long-tailed environments in a 
 \emph{Correction} $\rightarrow$ \emph{Modification} $\rightarrow$ \emph{Balance} learning manner.
On two newly proposed benchmarks, PLT-COCO and PLT-VOC, we demonstrate that the proposed framework significantly
outperforms existing MLC, LT-MLC and PL-MLC approaches. 
Our proposed PLT-MLC should serve as a complement to existing literature, providing new insights, and \method{} should serve as the first robust end-to-end baseline for the PLT-MLC problem.

{\small
\bibliographystyle{ieee_fullname}
\bibliography{egbib}
}

\end{document}